\documentclass[conference]{IEEEtran}
\IEEEoverridecommandlockouts
\usepackage{algorithmic}
\usepackage{graphicx}
\usepackage{xparse}
\usepackage[many]{tcolorbox}
\usepackage{xcolor}
\usepackage{mdframed}
\usepackage{csquotes}
\usepackage{amsmath}
\usepackage{amsthm}
\usepackage{amsfonts}
\usepackage{siunitx}
\usepackage{mathtools}
\usepackage{caption}
\usepackage{tabu}
\usepackage{multirow, makecell}
\usepackage{booktabs}
\usepackage{rotating}
\usepackage{tabularx}
\usepackage{array}
\usepackage{url}
\usepackage{adjustbox}

\usepackage{hyperref}

\newcommand{\eg}{\emph{e.g}. }
\newcommand{\ie}{\emph{i.e}. }

\newcommand\T{\rule{0pt}{2.9ex}}       % Top strut
\newcommand\B{\rule[-1.2ex]{0pt}{0pt}} % Bottom strut

\newcommand\mdoubleplus{\ensuremath{\mathbin{+\mkern-10mu+}}}

\graphicspath{{./images/pdf/}}

\newmdenv[
    topline=false,
    bottomline=false,
    rightline=false,
    linewidth=1pt,
    linecolor=gray,
    innerleftmargin=2pt,
    leftmargin=2pt,
    rightmargin=0pt,
    innertopmargin=0pt,
    innerbottommargin=0pt,
    innerrightmargin=0pt,
]{textbox}

\newcolumntype{Y}{>{\centering\arraybackslash}X}
\newcommand\RotTextNinety[1]{\rotatebox[origin=l]{90}{\parbox{1.7cm}{\centering#1}}}

\begin{document}

\title{Rank over Class: The Untapped Potential of Ranking in Natural Language Processing}

\author{\IEEEauthorblockN{Amir Atapour-Abarghouei}
\IEEEauthorblockA{\textit{Department of Computer Science} \\
\textit{Durham University}\\
Durham, UK \\
amir.atapour-abarghouei@durham.ac.uk}
\and
\IEEEauthorblockN{Stephen Bonner}
\IEEEauthorblockA{\textit{School of Computing} \\
\textit{Newcastle University}\\
Newcastle, UK \\
stephen.bonner3@newcastle.ac.uk}
\and
\IEEEauthorblockN{Andrew Stephen McGough}
\IEEEauthorblockA{\textit{School of Computing} \\
\textit{Newcastle University}\\
Newcastle, UK \\
stephen.mcgough@newcastle.ac.uk}
}

\maketitle

\begin{abstract}

    Text classification has long been a staple within Natural Language Processing (NLP) with applications spanning across diverse areas such as sentiment analysis, recommender systems and spam detection. With such a powerful solution, it is often tempting to use it as the go-to tool for all NLP problems since when you are holding a hammer, everything looks like a nail. However, we argue here that many tasks which are currently addressed using classification are in fact being shoehorned into a classification mould and that if we instead address them as a ranking problem, we not only improve the model, but we achieve better performance. We propose a novel end-to-end ranking approach consisting of a Transformer network responsible for producing representations for a pair of text sequences, which are in turn passed into a context aggregating network outputting ranking scores used to determine an ordering to the sequences based on some notion of relevance. We perform numerous experiments on publicly-available datasets and investigate the applications of ranking in problems often solved using classification. In an experiment on a heavily-skewed sentiment analysis dataset, converting ranking results to classification labels yields an approximately 22\% improvement over state-of-the-art text classification, demonstrating the efficacy of text ranking over text classification in certain scenarios.

\end{abstract}


\section{Introduction}

\label{sec:intro}

Recent advances in machine learning have led to significant strides in various active areas of research, with many applications already integrated into our daily lives. However, as natural languages are the primary method of human communication, learning-based Natural Language Processing (NLP) is receiving an ever-increasing level of attention within both academia and industry. Amongst the various applications of NLP, text classification \cite{kowsari2019text,minaee2020deep} has arguably blazed the trail due to the simplicity of its definition and its numerous use cases. From online content tagging \cite{salminen2019machine} to sentiment analysis \cite{zhang2018deep}, text classification has always been at the forefront of natural language processing.

With the emergence of deep learning, various approaches have addressed text classification via feed-forward networks using bag-of-word inputs \cite{iyyer2015deep}, recurrent neural networks that consider structural elements of the text \cite{wan2016deep} and convolutional neural networks capable of detecting position-invariant patterns in the text \cite{wang2017combining}. Transformers \cite{vaswani2017attention}, however, have arguably been the greatest advancement in NLP with significant improvements enabled by large-scale pre-trained language models, taking advantage of deeper architectures and larger corpora of text used for training to enable a stronger and better representation learning process. 

However, despite the significant advances in text classification, the field is still fraught with challenges. For instance, text classification can be highly subjective due to the presence of textual ambiguity, biased annotations and unknown classes. Examples of this are widely seen in the numerous publicly-available movie review datasets \cite{moviereview1,moviereview2,moviereview3}, commonly used as benchmarks for text classification. As an example, imagine the following sentence from a movie review:

\begin{textbox}
    \textbf{Example 1:} Movie Review - \emph{while plagued with a plodding mess of a narrative, the sincere performance of the main character salvages the clichéd dialogue and provides some escapism from the distorted perspective of the protagonist.}
\end{textbox}\vspace{-6pt}

Any observer would have trouble labelling this review as positive or negative. In the same vein, the performance of a machine learning model trained on subjectively annotated reviews would solely depend on the presence of similar words, patterns and structures of the text in other less ambiguous and more concrete data points in the dataset. This level of subjectivity and ambiguity essentially makes a fair and accurate classification of the sentiment of such passages very difficult using learning-based models.

Another significant issue in text classification stems from significant imbalances commonly found in existing datasets. If opinions on a specific topic are extracted from social media to be used as training data, most would lie on the extreme ends of the spectrum as individuals with extreme beliefs are more likely to voice their opinions publicly, thus creating a skew in the dataset towards more extreme opinions \cite{twitextreme}. In other scenarios, if subjective passages are labelled by multiple annotators with an averaging method determining the final label, most of the passages will inevitably fall near the centre of the labelling interval due to the subjectivity of the annotation \cite{rodrigues2014sequence}. Such imbalances can impede accurate classification and necessitate significant algorithmic intervention.

There are numerous other similar shortcomings in text classification, depending on the scope of the problem, the nature of the classes and the features available in the dataset. Here, we argue that many such issues can be resolved by reformulating certain text classification problems as \emph{text ranking}.
It is important to note that we are in no way suggesting that ranking should completely replace classification. On the contrary, classification remains an integral solution to a vast portion of the problems in computational linguistics. Our objective here is to investigate the possibility of reformulating certain key problems as ranking to address the common challenges hindering classification solutions.

Ranking is defined as a derivation of ordering over a list of items that maximises the utility of the entire list \cite{schutze2008introduction}. It is a significant component of many information retrieval systems with applications including web search, recommender systems, document summarisation and question answering \cite{li2011learning}.

Ranking, by nature, is different from classification and regression in that a classifier or a regressor attempts to assign a specific class label or value to an individual data point, while the objective of a ranking approach is to optimally sort a list, such that, for some notion of \emph{relevance}, the items within the list with higher relevance scores appear earlier in the list.

Note that while we primarily discuss the efficacy of ranking compared to classification in certain problems, the same conclusions can be made about regression solutions. The reason why classification is chosen as the main point of comparison is that the existing literature is dominated by text classification solutions to sentiment analysis \cite{zulqarnain2020comparative,minaee2021deep,joulin2016bag,mouthami2013sentiment} due to the availability of discrete class labels in benchmark datasets \cite{moviereview1,moviereview2,moviereview3} and the difficulty of attaining accurate continuous regression labels for subjective text data.

In this paper, we argue that for certain problems, ranking is a more appropriate formulation than classification (or regression). Take the movie review from \textit{Example 1}. While it might be difficult for a human or a neural network to accurately classify this as a positive or a negative review or associate it with an absolute sentiment value or class, it is significantly easier to rank this review with respect to other reviews as \emph{more positive} or \emph{more negative}, especially if they are placed in some known shared \emph{context}, for example all the reviews given for the same movie. In this paper, we explore the capabilities of text ranking and demonstrate its efficacy using experiments conducted over publicly-available datasets.

In short, the main contributions of this work are as follows:

\begin{itemize}
  \item With the end-to-end training of a Transformer, a context aggregating dense network and a pair-wise ranking loss, text passages are accurately ranked based on a given ranking label or score (Section \ref{subsec:approach:model}). %\vspace{-0.2cm}

  \item The potential of text ranking is demonstrated using various datasets \cite{moviereview3, StackExchange} with the learning process \emph{pivoting} around the different contexts the passages can be placed in, \eg the writer of a given passage or the topic of the passage, and the effectiveness of this contextual ranking is explored (Sections \ref{subsec:experiments:stack} and \ref{subsec:experiments:finefoods}). %\vspace{-0.2cm}
  
  \item When ranking results are artificially converted to class labels, our approach is capable of outperforming state-of-the-art text classification models on heavily-skewed text classification datasets \cite{Twitter} (Section \ref{subsec:experiments:twitter}). %\vspace{-0.2cm}
\end{itemize}

To enable reproducibility, an implementation of our ranking approach is publicly available\footnote{\href{https://github.com/atapour/rank-over-class}{https://github.com/atapour/rank-over-class}}. %\vspace{-0.05cm}

% %-------------------------------------------------------------------------
% %--------------------------- END OF SECTION ------------------------------
% %-------------------------------------------------------------------------

% %-------------------------------------------------------------------------
% %----------------------------- NEW SECTION -------------------------------
% %-------------------------------------------------------------------------
\section{Related Work}
\label{sec:related_work}

Having first appeared in the literature in the 1940s \cite{leontief1941structure}, ranking gained prominence as the foundation of modern search engines towards the end of the last millennium \cite{page1999pagerank}. Given a query $q$ and a collection $P$ of passages $p$ that match the query, the goal is to rank all the passages $p$ in $P$ according to some notion of relevance to $q$ so that the \emph{best} results appear earlier in the ranked list. Note that while we, in this work, focus on passage ranking, the same concepts can equally be applied to any data modality.

Though traditionally solved via boolean, vector space and probabilistic models \cite{jones2000probabilistic,baeza1999modern}, ranking is now commonly performed using learning-based approaches \cite{li2011learning}, taking advantage of labelled data and some parametrised function to map feature vectors extracted from list items to real values used as ranking scores. This function is subsequently used to sort the items. Ranking approaches can be point-wise, pair-wise or list-wise.

Point-wise approaches \cite{gey1994inferring} utilise a classifier or a regressor trained to predict the relevance score of a passage, which is subsequently used to sort the items, with respect to a given query. Pair-wise techniques \cite{burges2010ranknet}, on the other hand, consider a pair of items at each pass with the overall objective being to minimise the number of inversions, where the items are in the wrong order relative to the ground truth, in the ranked list. List-wise approaches \cite{wang2018lambdaloss} attempt to solve for the optimal ranking for the entire list all at once. This is often done either via a specific loss function design, based on the unique properties of the items \cite{lan2014position}, or by directly optimising certain information retrieval metrics \cite{taylor2008softrank}.

In this paper, we propose a pair-wise ranking approach, capable of outperforming classification in certain natural language processing problems. Unlike point-wise approaches, a pair-wise technique, such as the one proposed here, receives more context as it directly compares list items. List-wise models, despite considering the explicit global context and removing the need for post-processing, are more difficult to optimise, slower to converge and suffer from training instability issues, which is why a pair-wise ranking approach is implemented for this study.
%-------------------------------------------------------------------------
%--------------------------- END OF SECTION ------------------------------
%-------------------------------------------------------------------------

%...............................
\begin{figure*}[t!]
	\centering
	\includegraphics[width=0.8\linewidth]{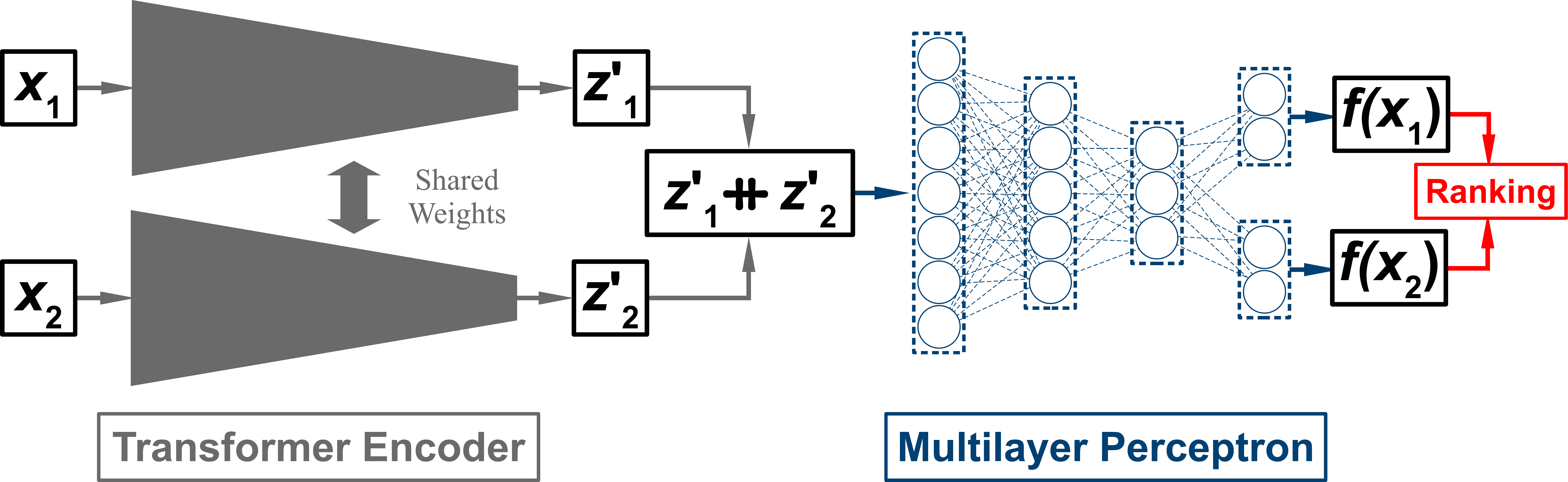}
	\captionsetup[figure]{skip=7pt}
	\captionof{figure}{The overall model approximates the ranking function, $f$. Input passages, $x_1$ and $x_2$, are first passed through a Transformer network. A multilayer perceptron then takes the concatenated Transformer outputs ($z^\prime_1$ $\mdoubleplus$ $z^\prime_2$) and produces the ranking scores, $f(x_1)$ and $f(x_2)$, used to rank the inputs.}
	\label{fig:pipeline}\vspace{-0.25cm}
\end{figure*}
%...............................

%-------------------------------------------------------------------------
%----------------------------- NEW SECTION -------------------------------
%-------------------------------------------------------------------------
\section{Proposed Approach}
\label{sec:approach}

Consisting of two sub-networks trained end-to-end, our overall approach receives two passages as its input and outputs two scalar values which can be used to determine whether the passages have been input in the correct order. See Figure \ref{fig:pipeline}.

The first sub-network (left) is a Transformer that produces a representation of the input passages. The second sub-network (right) is a context aggregating multilayer perceptron, which receives the latent vectors representing the passages (produced by the Transformer network) and outputs ranking scores, used to determine the ordering of the two input passages.

The proposed approach performs contextually sound comparisons between passage pairs to ensure meaningful ranking results. Details of the approach are discussed in the following.
%--------------------------- END OF SUBSECTION ------------------------------

%----------------------------- NEW SUBSECTION -------------------------------
\subsection{Context is King}
\label{subsec:approach:model}

In text classification, either it can be assumed that the entirety of the training dataset falls within the same context or the context is completely ignored. Context, in this setting, refers to a unifying element which the passages can be grouped by and provides a coherent background for understanding what those passages represent. To get a clearer picture of what is meant by context, let's refer to the following examples:

\begin{textbox}
    \textbf{Example 2:} a human observer is given three random passages and asked to assess their quality. One passage is an excerpt from a technical report, one a poem and the other from a history textbook.
\end{textbox}
\vspace{-6pt}
\begin{textbox}
    \textbf{Example 3:} a human observer is given three random passages and asked to assess their quality. All three passages contain answers from different individuals to the same question.
\end{textbox}\vspace{-6pt}

In \emph{Example 2}, the passages are from different sources, follow different linguistic patterns and address different types of information. As such, they lack a shared context. This makes objectively comparing their quality extremely difficult, even for a human expert. However, in \emph{Example 3}, the passages are all answers to the same question and follow a shared context. This means the problem can be solved in a more meaningful and objective manner. In this vein, our approach first attempts to introduce the idea of \emph{context}.

To bring the notion of \emph{context} of the passages into the learning process, we first group the passages in our dataset based on some shared context, thereby pivoting the learning process around it. This provides a stronger background for the model's representation learning capabilities as the model, at any given point in time, is only being trained on passages from the same context. From now on, we will refer to this as the \emph{contextual pivot point}. Subsequently, once the passages are grouped based on a contextual pivot point, all possible \emph{combinations} of passages within the groups are extracted and used as training data for our pair-wise ranking model.

The overall pipeline of our approach is seen in Figure \ref{fig:pipeline}. A pair of input passages are passed through the layers of a headless Transformer network to get a latent vector containing the representation of each passage. In our experiments, BERT \cite{devlin2019bert}, GPT2 \cite{radford2019language}, RoBERTa \cite{liu2019roberta} and ALBERT v2 \cite{lan2019albert} are used to obtain the feature vectors but any other Transformer model can similarly be used. As seen in Figure \ref{fig:pipeline}, the resulting feature vectors of the two passages are concatenated and used as the input to a four-layered multilayer perceptron (context-aggregating network). The multilayer perceptron assesses the relationship between the two inputs and regresses to two values (ranking scores), subsequently used to rank the passages.

Note that both the input passage pairs and the output score pairs are correspondingly-ordered and any change in the ordering of the input passages affects the output ranking scores. However, this does not mean that the results will be inconsistent when the input changes, but that the interpretation of the output ranking scores depends on the order of the input passages (similar to a comparison operator). Trained end to end, the entire model uses a ranking loss function (explained in Section \ref{subsec:approach:loss}) and pair-wise ranking is enabled via the comparison of the output values.
%--------------------------- END OF SUBSECTION ------------------------------

%............................................................................
\begin{table*}[!t]
	\centering
	\resizebox{0.78\linewidth}{!}{
		{\tabulinesep=0mm
			\begin{tabu}{@{\extracolsep{12pt}}l c c c c c c@{}}
				\hline\hline
				\multicolumn{1}{l}{\multirow{2}{*}{Classification Model}} & 
				\multicolumn{3}{c}{(a) Stack Exchange} &
				\multicolumn{3}{c}{(b) Fine Foods}\T\B \\
				\cline{2-4} \cline{5-7}
						& Accuracy	& F\textsubscript{1} Score & AUC & Accuracy	& F\textsubscript{1} Score & AUC\T\B\\
				\hline\hline
ALBERT \cite{lan2019albert}				& 0.272 		& 0.278 & 0.618 & 	0.722 		& 0.617		& 0.810\T\\
RoBERTa	\cite{liu2019roberta}			& 0.281 		& 0.291 & 0.635	& 	0.705 		& 0.590		& 0.809\\
GPT2 \cite{radford2019language}			& 0.285 		& 0.298 & 0.622	& 	0.765 		& 0.738		& 0.813\\
BERT \cite{devlin2019bert}				& 0.292 		& 0.309 & 0.648	&	0.768 		& 0.721		& 0.828\B\\

\hline
\hline
        \end{tabu}
    }
}
\captionsetup[table]{skip=7pt}
\captionof{table}{(a) Quality assessment of Stack Exchange posts and (b) sentiment analysis of Fine Food Reviews.}
\label{table:stack_foods_classification}
\end{table*}
%............................................................................

%----------------------------- NEW SUBSECTION -------------------------------
\subsection{Loss Function}
\label{subsec:approach:loss}

While most ranking approaches traditionally utilise loss functions such as the sigmoid cross-entropy for binary relevance labels, pair-wise logistic loss or softmax cross-entropy \cite{pasumarthi2019tf}, we make use of a margin-based ranking loss, which has been effectively used for representation learning in knowledge graph embedding models \cite{trouillon2016complex} by separating the positive samples from the negative samples within the dataset by a given margin. Using this loss function, we can take advantage of its strong representation learning capabilities to extract more robust features from the passages and better learn their compositional relationship during ranking. Formally, for any passage pair $p_i$ and $p_j$, the ranking label, $E$, is determined as:

\begin{equation}
	\begin{adjustbox}{max width=185pt}
	$
		E(p_i, p_j)=\begin{cases}
		1,  &  p_i \text{ ranked higher than } p_j\\
		\text{-}1, &  p_j \text{ ranked higher than } p_i
		\end{cases},
	\label{eq:label_E}
	$
	\end{adjustbox}
\end{equation}\vspace{0.2cm}
\\
where $E$ is the ground truth ranking label. Consequently, for the set of all passage pairs $\Psi = \{(p_i, p_j) ; E(p_i, p_j)\}$, a pair is fed into our model, which approximates the desired ranking function $f$. The loss function is thus defined as:

\begin{equation}
	\begin{adjustbox}{max width=195pt}
	$
	\mathcal{L} = \smashoperator{\sum_{(p_i, p_j) \in \Psi}} \max(0, -E(p_i, p_j) \times (f(p_i) - f(p_j)) + \gamma),
		\label{eq:sum-depth-overall}
	$
	\end{adjustbox}
\end{equation}\vspace{0.2cm}
\\
where $f(p_i)$ and $f(p_j)$ are the ranking scores produced by the overall model, $f$, and $\gamma$ is the margin enforced between $f(p_i)$ and $f(p_j)$. In our experiments, $\gamma=2$ empirically yields the most favourable results. This loss measures ranking violations of passage pairs, allowing the network to learn discriminative features to enforce a clearer distinction between the passages.

%--------------------------- END OF SUBSECTION ------------------------------

%----------------------------- NEW SUBSECTION -------------------------------
\subsection{Implementation Details}
\label{subsec:approach:implementation}

For all experiments, the smallest pre-trained versions of the Transformers provided by the HuggingFace library \cite{wolf2019transformers} are used. Text sequences are truncated if they exceed 128 tokens, except for experiments in Section \ref{subsec:experiments:msmarco}, where sequences of 512 tokens are used. The multilayer perceptron uses Linear-BatchNorm-PReLu modules with a dropout of 0.2 for each layer during training. All implementations are done in PyTorch \cite{pytorch}, with AdamW \cite{loshchilov2017decoupled} providing the optimization ($\beta_{1} = 0.9$, $\beta_{2} = 0.999$, $\epsilon=$1e-8).

An advantage of our approach is that there are very few hyperparameters associated with it. The only hyperparameters that might have an effect on the overall results of the approach in the experiments are the margin value in the loss function ($\gamma$) and the optimisation parameters (\eg learning rate). These hyperparameters are empirically selected through a grid search. For our hyperparameter searches, 10\% of our Stack Exchange dataset (explained in Section \ref{subsec:experiments:stack}) is randomly selected as the training dataset and 2\% as the test set. The margin value $\gamma$ is selected from values of \{0, 0.1, 1, 2, 5, 10\} and the learning rate $\alpha$ is selected from \{1e-6, 4e-6, 8e-6, 1e-5, 4e-5, 8e-5, 1e-4, 4e-4, 8e-4\}. Our experiments indicate that the best results are obtained when the margin value is $\gamma = 2$ for all models and the learning rate is $\alpha = $4e-5 for BERT and GPT2 and $\alpha = 4$e-6 for ALBERT and RoBERTa.

All the Transformer networks used in our work are trained from a fixed pre-trained state and the only portion of our overall models that actually starts from random initialisation is the four-layer context aggregating fully-connected network. This greatly reduces the effect of random initialisation on our approach as the size of the context aggregating multilayer perceptron pales in comparison to the size of the Transformer networks.

All experiments are carried out using two \href{https://www.nvidia.com/en-gb/deep-learning-ai/products/titan-rtx/}{NVIDIA Titan RTX} GPUs in parallel with a combined memory of 48 GB on an \href{https://www.archlinux.org/}{Arch Linux} system with a 3.30GHz 10-core Intel CPU and 64 GB of memory. Note that for large NLP models, such as those used in this work, this hardware has limited capabilities, which is why smaller Transformer models and datasets are used in the approach.
% -------------------------------------------------------------------------
% --------------------------- END OF SECTION ------------------------------
%-------------------------------------------------------------------------

%-------------------------------------------------------------------------
%----------------------------- NEW SECTION -------------------------------
%-------------------------------------------------------------------------
\section{Experimental Results}
\label{sec:experiments}

To evaluate our approach, we perform extensive experiments using four publicly-available datasets, Stack Exchange \cite{StackExchange}, Fine Food Reviews \cite{moviereview3}, tweets about self-driving cars \cite{Twitter} and the MS MARCO passage ranking dataset \cite{bajaj2016ms}. The datasets are selected for their level of difficulty and prevalence in the sentiment analysis and text classification literature. 

Note that the tweets about self-driving cars are downloaded from an anonymised publicly available dataset \cite{Twitter} and not extracted by the authors of this paper. In addition to the experiments conducted to assess the efficacy of our technique, we also perform ablation studies and demonstrate the importance of every component of the proposed approach.

%----------------------------- NEW SUBSECTION -------------------------------
\subsection{Stack Exchange}
\label{subsec:experiments:stack}

A large Question and Answer Forum made up of numerous communities focusing on different subjects, Stack Exchange \cite{StackExchange} offers a space where users can post questions about specific topics for which they can receive answers from other users. Any user can annotate whether an answer is useful or not by voting for it favourably (up-vote) or unfavourably (down-vote). The original asker can also mark one answer as the best. As these votes are expected to be based on quality, subjective though they might be, they provide an opportunity for a learning-based approach to assess the quality of each answer for a potential recommendation system.

A na\"ive way to assess post quality (in this work only the \emph{answers} are considered for quality assessment in the training and testing datasets) is using a classification-based solution that predicts the quality of posts using classes corresponding to how well the post has done (or can do) in terms of up-votes. The ground truth quality scores can be generated based on the number of votes an answer has received normalised by the number of votes the original question has received.

This normalisation is important since questions with higher vote counts generally receive more engagement from the users. As such, the normalisation process removes the possibility of bad answers to more popular questions with a higher vote count overwhelming better answers to less popular questions. The quality scores are then categorised into 5 classes via a simple histogram.

The dataset is extracted from the communities of \emph{Ask Ubuntu}, \emph{Cryptography}, \emph{Data Science}, \emph{Network Engineering}, \emph{Unix \& Linux} and \emph{Webmasters}. 250,000 posts are randomly selected for the training dataset and 50,000 for testing. State-of-the-art text classification models \cite{devlin2019bert,radford2019language,liu2019roberta,lan2019albert} are trained for 3 epochs (to convergence) to classify the posts based on their quality scores.

%...............................
\begin{figure*}[t!]
	\centering
	\includegraphics[width=0.7\linewidth]{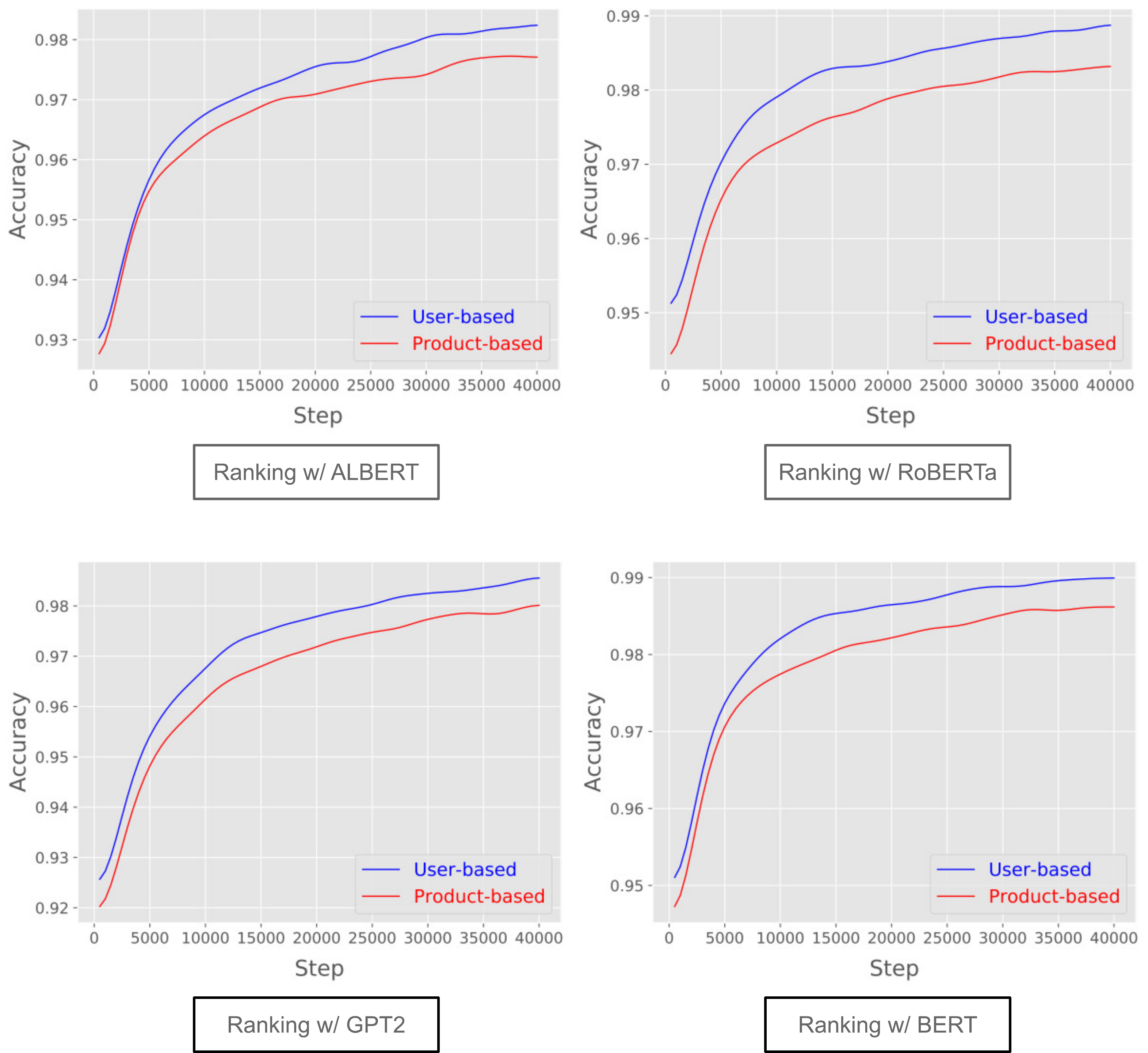}
	\captionsetup[figure]{skip=7pt}
	\captionof{figure}{Test accuracy of pair rankings on the Fine Foods reviews dataset using our model trained for 40,000 steps. When ranking is pivoted around users, the models reach convergence faster, due to the overpowering effect of user writing styles.}
	\label{fig:plots}
\end{figure*}
%...............................

%............................................................................
\begin{table}[!t]
	\centering
	\resizebox{\linewidth}{!}{
		{\tabulinesep=0mm
		\begin{tabu}{@{\extracolsep{2pt}}p{0.62cm} l c c c c@{}}

			\hline\hline
			\multicolumn{1}{c}{\multirow{2}{*}{}} & 
			\multicolumn{1}{c}{\multirow{2}{*}{Approach}} & 
			\multicolumn{3}{c}{Ranking Metrics @$k$} &
			\multicolumn{1}{c}{Pair Labels}\T\B\\

			\cline{3-5} \cline{6-6} 

			\multicolumn{1}{c}{}    & & MRR	& NDCG    & MAP & Accuracy\T\B\\
			\hline\hline

\multirow{4}{*}[2.2pt]{\RotTextNinety{(a) Stack Exchange}} & 
\multicolumn{1}{|l}{w/ ALBERT}	 	& 0.656 & 0.772 & 0.662 & 0.854\T\\
& \multicolumn{1}{|l}{w/ RoBERTa}	& 0.671 & 0.780 & 0.678 & 0.882\\
& \multicolumn{1}{|l}{w/ GPT2}	 	& 0.760 & 0.806 & 0.764 & 0.895\\
& \multicolumn{1}{|l}{w/ BERT}	 	& \textbf{0.762} & \textbf{0.814} & \textbf{0.768} & \textbf{0.910}\B\\

\hline

\multirow{4}{*}[2.2pt]{\RotTextNinety{(b) Review (User)}} & 
\multicolumn{1}{|l}{w/ ALBERT}		& 0.970 & 0.982 & 0.976 & 0.985\T\\
& \multicolumn{1}{|l}{w/ RoBERTa}	& 0.972 & 0.984 & 0.979 & 0.990\\
& \multicolumn{1}{|l}{w/ GPT2}		& 0.975 & \textbf{0.986} & 0.982 & \textbf{0.991}\\
& \multicolumn{1}{|l}{w/ BERT}		& \textbf{0.977} & \textbf{0.986}	& \textbf{0.984} & 0.990\B\\

\hline

\multirow{4}{*}[2.2pt]{\RotTextNinety{(c) Review (Product)}} & 
\multicolumn{1}{|l}{w/ ALBERT}	 	& 0.972 & 0.984 & 0.976 & 0.981\T\\
& \multicolumn{1}{|l}{w/ RoBERTa}	& 0.970 & 0.982 & 0.972 & 0.985\\
& \multicolumn{1}{|l}{w/ GPT2}	 	& \textbf{0.976} & \textbf{0.987} & \textbf{0.982} & \textbf{0.990}\\
& \multicolumn{1}{|l}{w/ BERT}	 	& 0.974 & 0.985 & 0.981 & \textbf{0.990}\B\\

\hline
\hline

        \end{tabu}
    }
}
\captionsetup[table]{skip=7pt}
\captionof{table}{Results of passage ranking using the proposed approach with data from (a) Stack Exchange, (b) Food Reviews with users as the contextual pivot point and (c) Food Reviews with products as the pivot point.}
\label{table:stack_foods_tweets_ranking}
\end{table}
%............................................................................

As seen in Table \ref{table:stack_foods_classification} (a), all classification models fail to assess the quality of the posts beyond randomly guessing, evidenced by the low accuracy, F\textsubscript{1} Score and Area Under the ROC Curve (AUC). This is explained by the lack of context providing a coherent backdrop for the passages as they are associated with different users, questions and communities. Note that while in this paper, we focus on classification as a commonly used baseline, experiments with a regression-based models, predicting the quality score of the passages directly, demonstrated similarly bad performance, explained by the extreme sensitivity of regression models to data imbalances.

Using \textit{Example 2} in Section \ref{subsec:approach:model}, we explained how even a human would have trouble assessing the text quality of unrelated passages without any shared context. The same concept applies to this dataset, where even a human observer would have difficulty comparing the quality of answers written by different people to different questions. This is why a contextual pivot point is necessary.

The Stack Exchange post quality annotations (user votes) are within the context of each question, \ie users vote on post quality based on how well the post addresses the specific question. Therefore, measuring the quality of the posts within each question would be very easy as the desired task directly aligns with the ground truth data. As a result, we experiment with unique users as the contextual pivot point. There are 17,085 unique users within our dataset with an average of 20.1 answers per user.

%............................................................................
\begin{table}[!t]
	\centering
	\resizebox{0.78\columnwidth}{!}{
		{\tabulinesep=0mm
			\begin{tabu}{@{\extracolsep{5pt}}l c c@{}}
				\hline\hline
				\multicolumn{1}{l}{\multirow{2}{*}{Ranking Model}} & 
				\multicolumn{2}{c}{Pair Labels}\T\B\\
				\cline{2-3}
						& Accuracy	& F\textsubscript{1} Score\T\B\\
				\hline\hline
Ranking w/ ALBERT				& 0.905 			& 0.908\T\\
Ranking w/ RoBERTa				& 0.910 			& 0.917\\
Ranking w/ GPT2					& 0.932 			& 0.936\\
Ranking w/ BERT					& 0.945 			& 0.950\B\\

\hline
\hline
        \end{tabu}
    }
}
\captionsetup[table]{skip=7pt}
\captionof{table}{Ranking results of the proposed approach over 1,633 comparisons of temporally-ordered passages from 50 users with at least 20 answers that have unique scores \cite{StackExchange}.}
\label{table:stack_time_pattern}
\end{table}
%............................................................................

In our experiments, we assess the quality of passages posted by individual users by ranking them. While evaluating the quality of posts for \emph{each individual user} is possible via text classification, since classification cannot explicitly take context into account, it would entail training \emph{a separate classification model for each individual user} (17,085 different classification models for the 17,085 users), which is intractable. Our ranking approach, however, is trained end to end over the entire dataset once and can provide an accurate measure of the quality for the posts across all questions, all users and all communities.

With the user chosen as the contextual pivot point, the passages are grouped based on unique users and all combinations of passages are extracted as training data ($ \approx$8,000,000 pairs). Similar to the classification setup, ground truth ranking labels (Eqn. \ref{eq:label_E}) are based on the number of votes an answer has received normalised by the number of votes the original question has received. To enable evaluation with the appropriate ranking metrics, 100 random users with at least 100 answers that have unique quality scores are selected as the test set.

Table \ref{table:stack_foods_tweets_ranking} (a) demonstrates better results are obtained with BERT and GPT2. Common ranking metrics, including Normalized Discounted Cumulative Gain (NDCG), Mean Average Precision (MAP) and Mean Reciprocal Rank (MRR) are used to evaluate the results with $k=10$. Note that $k$ denotes how many items are considered relevant in the ranked list - \ie relevance is only considered in the top $k$ items in the list for the ranking metrics.

We also measure the accuracy of the approach applied to test passages (with success referring to an accurate ranking of pairs - Eqn. \ref{eq:label_E}). As seen in Table \ref{table:stack_foods_tweets_ranking} (a), despite the fact that the ground truth labels are crowd-sourced and not objective, accurate ranking is achieved.

An interesting application of ranking the passages with the contextual pivot point being the users is the possibility of tracking the change in the skill level of users over time. To test this, we extract a test set of 50 random users with at least 20 answers that have unique ranking scores. Considering that the model receives no information about the time of the posts, if the resulting ranking matches the order at which passages were posted, the approach is capable of tracking the skill level of users, which can be of value in downstream recommendation and forum moderation systems, with opportunities for recommending questions to users and ordering posts based on a user's skill level, among others.

A numerical analysis of this is shown in Table \ref{table:stack_time_pattern}, which presents the results after evaluating the ranking of posts from 50 users. Since for this experiment, the accuracy of the comparison of the pairs is what matters, we present the accuracy metric over the 1,633 comparisons of temporally-ordered passage pairs. The promising results in Table \ref{table:stack_time_pattern} are indicative of the potential for better user behaviour and profile tracking, leading to more accurate recommendation and security systems.

We would like to also point out that while we demonstrate how classification is not a suitable solution for this problem (Table \ref{table:stack_foods_classification} (a)), a regression-based solution also suffers from the same issues. The quality scores for the Stack Overflow posts are segmented into 5 classes via a simple histogram. This leads to an unbalanced dataset as many of the posts are of roughly the same quality with a number of outliers with significantly different scores. These issues will remain the same in a regression setup with most posts having the same score and the outliers skewing the possible range of scores the model is meant to learn, leading to potentially worse results than classification in general. This is why the literature on quality assessment/sentiment analysis predominantly focuses on classification solutions \cite{ormandjieva2007toward,ganguly2014automatic,haque2018sentiment,johnson2017deep}.

Due to the fact that the majority of the posts are clumped together in terms of their score and the existence of outliers, a regression model completely fails to capture the underlying data distribution and will perform no better than classification. Our ranking model, however, does not suffer from issues stemming from data imbalance, better captures the patterns and the associations in the posts and their quality and produces promising results, as seen in Tables \ref{table:stack_foods_tweets_ranking} (a) and \ref{table:stack_time_pattern}.

%--------------------------- END OF SUBSECTION ------------------------------

%----------------------------- NEW SUBSECTION -------------------------------
\subsection{Fine Food Reviews}
\label{subsec:experiments:finefoods}

The Amazon Fine Food Reviews dataset \cite{moviereview3} contains approximately 500,000 reviews of around 75,000 products from roughly 250,000 users. The objective is to predict the sentiment of the review based on scores (from 1 to 5) provided in the dataset. As opposed to the Stack Exchange data, the passages in this dataset are significantly easier to classify as they substantially follow the same context (product reviews).

Experimental results presented in Table \ref{table:stack_foods_classification} (b) demonstrate that classification methods are reasonably capable of classifying the reviews albeit with underwhelming results. 20\% of the data is randomly selected as the test set and the text sequences are truncated to 128 tokens (words). The dataset is unbalanced with higher sentiment scores being more prevalent in the dataset. However, if the task of sentiment analysis is reformulated as ranking, significantly better performance can be expected.

For the purposes of our approach, the contextual pivot point for this dataset can be either the \textit{user} (ranking the sentiment of the reviews from individual users for all the products they have reviewed) or the \textit{product} (ranking the sentiment of the reviews of individual products from all users who have reviewed them). Here, we perform experiments based on both the users and the products as contextual pivot points.

Separate training and testing sets are created, with the product-based test set made up of 200 random products with reviews that have unique scores (only 5 unique scores exist) and the user-based test set made up of 200 random users with reviews that have unique scores. To evaluate ranking, the test reviews for each user or product are selected not to have the same score so the ranking of the ground truth list for each user/product can be easily compared against the ranking output of the approach. As our model always ranks a pair of passages, even if they are similar, only pairs with different ranking scores are passed into the model during training and in the test set, only passages with unique scores are used to evaluate the model. 

Table \ref{table:stack_foods_tweets_ranking} (b) shows the results of the user-based experiments and Table \ref{table:stack_foods_tweets_ranking} (c) the results of the product-based experiments. As there are only 5 items for each user or product (5 sentiment classes), the ranking metrics are calculated with $k=2$. \emph{Accuracy} alludes to the correctness of pair rankings over all possible pairs in the test set. The metric values are higher than those of Section \ref{subsec:experiments:stack}, as only five items exist in the list, but as seen in Table \ref{table:stack_foods_tweets_ranking}, the results are extremely promising with the models achieving near perfect ranking in both experiments.

While highly accurate results are achieved across the board, an interesting observation is that it is easier for the model to learn the context and thus perform better ranking when the pairs are pivoted around users as opposed to products. Figure \ref{fig:plots} demonstrates how all models consistently reach convergence faster when the contextual pivot point is the user. This is due to the powerful influence of the writing style of users with different reviews. While the reviews of all products share a clear context as evidenced by the results in Table \ref{table:stack_foods_tweets_ranking} (c), user-based ranking is more easily learned (Figure \ref{fig:plots}).
%--------------------------- END OF SUBSECTION ------------------------------

%----------------------------- NEW SUBSECTION -------------------------------
\subsection{Sentiment Analysis: Self-Driving Cars}
\label{subsec:experiments:twitter}

This dataset \cite{Twitter} focuses on people's opinions of autonomous driving and consists of 6,943 relevant tweets labelled from 1 to 5, with 1 indicating the most negative sentiment and 5 the most positive. This dataset is heavily skewed with over 61\% of the data annotated as \emph{neutral} (4,245 tweets with label 3) and less than 2\% annotated as \emph{very negative} (110 tweets with label 1). This creates significant challenges for any classification approach and requires extreme measures to combat the imbalance in the dataset. Note that this datasets is not directly obtained from Twitter by the authors of this work and is instead downloaded as an anonymised publicly-available dataset \cite{Twitter}.

%............................................................................
\begin{table}[!t]
	\centering
	\resizebox{\columnwidth}{!}{
		{\tabulinesep=0mm
			\begin{tabu}{@{\extracolsep{5pt}}l c c c@{}}
				\hline\hline
				\multicolumn{1}{l}{\multirow{2}{*}{Classification Model}} & 
				\multicolumn{3}{c}{Evaluation Metrics}\T\B\\
				\cline{2-4}
									& Accuracy	& F\textsubscript{1} Score & AUC\T\B\\
\hline\hline

Liu et al. \cite{liu2019roberta}				& 0.668 		& 0.598		& 0.642\T\\
Lan et al. \cite{lan2019albert}				& 0.672 		& 0.602		& 0.626\\
Radford et al. \cite{radford2019language}			& 0.684 		& 0.608		& 0.638\\
Devlin et al. \cite{devlin2019bert}				& 0.688	 		& 0.620		& 0.634\\
Yang et al. \cite{yang2019xlnet}				& 0.692 		& 0.626		& 0.642\B\\

\hline

Converted Ranking w/ ALBERT 		& 0.872 		& --		& --\T\\
Converted Ranking w/ RoBERTa 		& 0.902	 		& --		& --\\
Converted Ranking w/ GPT2 			& 0.898	 		& -- 		& --\\
Converted Ranking w/ BERT 			& 0.916 		& --	 	& --\B\\

\hline
\hline
        \end{tabu}
    }
}
\captionsetup[table]{skip=7pt}
\captionof{table}{Accuracy results of state-of-the-art classification compared against our ranking approach converted to class labels on the Self-Driving Car Tweets dataset \cite{Twitter}.}
\label{table:twitter}
\end{table}
%............................................................................

Here, we evaluate the ability of our model to deal with such a challenging dataset. While all the tweets follow the same context (self-driving vehicles), the user that posted each tweet is ignored in this dataset and thus no contextual pivot point is available other than the shared subject of the tweets (self-driving vehicles). The dataset is split into an unbalanced training set and a balanced test set with the test data containing 100 tweets from each class (500 tweets in total). This leaves only 10 tweets from class 1 for training, further exacerbating the data imbalance problem.

Note that the only reason that the test set for this experiment is balanced is to enable a better analysis of the results of our comparators. Our ranking approach is not affected by any imbalances in the training and/or test datasets and is able to remain completely unchanged even if the test set was extremely unbalanced.

This is intuitively expected as when one ranks a series of data points based on some score, the distribution of the data and the score does not matter as the ranking technique only needs to predict an ordering of the data and not their class/value or score -- each pair for comparison are handled independently and there is no longer a concept of balance to the data. For a classification or regression approach, however, distribution is of significant importance. We have opted for a balanced test set only to make the evaluation of the classification techniques easier.

As our approach is a ranking one, it is capable of producing an ordered list and does not predict absolute class labels. To enable a direct comparison with classification methods, here, we convert the resulting ranked list to a set of class labels, which can be trivial as the class labels represent sentiment scores from 5 to 1 (most positive to most negative) and our ranking approach is trained to rank based on sentiment scores (most positive to most negative). This is accomplished by simply sorting the balanced test set of 500 tweets from most positive to most negative using our ranking approach, dividing the ranked list into 5 segments and assigning labels (5 to 1) to each segment from top to bottom.

It is important to note that the balance or imbalance of the test set has no bearing on the performance of our approach. For example, if only 5\% of the test data belonged to class 5 (most positive) instead of the current 20\%, it would make no difference to our approach as those would still be ranked at the top of the sorted list. Ranking does not suffer from such imbalance issues, which is why we argue it is more suited for solving such problems than classification solutions.

Metrics such as F\textsubscript{1} and AUC do not apply to a ranking approach even when the results are converted to class labels. However, accuracy provides an excellent metric for evaluating the performance of the state-of-the-art classification methods as well as our ranking technique. So while we report accuracy, F\textsubscript{1} and AUC for classification methods, our converted ranking is evaluated using accuracy as the primary metric (Table \ref{table:twitter}).

%............................................................................
\begin{table}[!t]
	\centering
	\resizebox{0.65\columnwidth}{!}{
		{\tabulinesep=0mm
			\begin{tabu}{@{\extracolsep{5pt}}l c @{}}
				\hline\hline
				\multicolumn{1}{c}{\multirow{2}{*}{Ranking Approach}} & 
				\multicolumn{1}{c}{Dev}\T\B\\
				\cline{2-2}
						& MRR@10\T\B\\
				\hline\hline

Mitra and Craswell \cite{mitra2019updated}				& 0.252\T\\
Mitra et al. \cite{mitra2019incorporating}		& 0.254\\
Rosset et al. \cite{rosset2019axiomatic}			& 0.262\\
Nogueira et al. \cite{nogueira2019doc2query}		& 0.277\\
Qiao et al. \cite{qiao2019understanding}		& 0.311\\
Hofstatter et al.\cite{hofstatter2019effect}			& 0.318\B\\

\hline
Our Approach w/ ALBERT				& 0.264\T\\
Our Approach w/ RoBERTa				& 0.267\\
Our Approach w/ GPT2				& 0.273\\
Our Approach w/ BERT				& 0.275\B\\

\hline
\hline
        \end{tabu}
    }
}
\captionsetup[table]{skip=7pt}
\captionof{table}{Comparison of contemporary ranking methods and our approach applied to MS MARCO. It is important to note that unlike many comparators, the proposed approach uses the smallest most basic version of the Transformers.}
\label{table:msmarco}
\end{table}
%............................................................................

We would like to emphasise that this conversion of the ranking results to class labels is \emph{not} necessary in the real world and is only done to evaluate our approach in direct comparison to classification. We argue that in certain problems, such as the one addressed in this section, ranking is a better solution and assigning an absolute class or value to a passage does not offer as much information as an ordered list of passages with a shared context provides. 

In this experiment, all models are trained for 100,000 steps. Table \ref{table:twitter} shows how effective our approach is by achieving promising results despite the significant imbalance in the training dataset. Based on the F\textsubscript{1} and AUC scores of the state-of-the-art text classification methods \cite{liu2019roberta,lan2019albert,radford2019language,devlin2019bert,yang2019xlnet}, it is clear that the skew in the training dataset has severely affected the learning capabilities of the state-of-the-art classifiers.

On the other hand, ranking remains robust. In fact, the converted results of our best ranking approach improve on the most effective text classifier \cite{yang2019xlnet} by about 22\%.
%--------------------------- END OF SUBSECTION ------------------------------

%----------------------------- NEW SUBSECTION -------------------------------
\subsection{MS MARCO}
\label{subsec:experiments:msmarco}

While the primary focus of this paper is not information retrieval, ranking is indelibly linked to information retrieval so much so that the metrics used to evaluate our results are predominantly from the information retrieval literature. In this vein, we also apply our approach to the publicly available benchmark dataset of MS MARCO \cite{bajaj2016ms}.

We perform the passage ranking task within the benchmark using our proposed pipeline. The queries and the passages are concatenated into sequences of no more 512 tokens. Due to hardware restrictions, less than 1\% of the entire available dataset is used for training and the smallest possible versions of the Transformer models are used.

Table \ref{table:msmarco} demonstrates the results of our approach compared to contemporary ranking approaches on the validation set of the MS MARCO dataset. Despite using smaller Transformer models and a fraction of the available training data, our proposed approach produces promising results and remains competitive with dedicated information retrieval techniques.
%--------------------------- END OF SUBSECTION ------------------------------

%----------------------------- NEW SUBSECTION -------------------------------
\subsection{Ablation Studies}
\label{subsec:experiments:ablation}

It is important to evaluate and demonstrate the contribution of every component of our approach. Consequently, we re-train our model with different components removed or replaced, with pair ranking accuracy being measured as the primary metric. For these experiments, 30\% of the Stack Exchange dataset is selected for our ablation studies.

As an integral part of our pipeline is the Transformer model generating the feature vector representing the input sequences, the choice of the Transformer is a critical decision. We perform all our experiments with four commonly-used state-of-the-art Transformers. Table \ref{table:ablation} demonstrates that BERT produces superior results with GPT2 only remaining competitive despite its larger size. This is also supported by our other experiments with results presented in Tables \ref{table:stack_foods_tweets_ranking}, \ref{table:stack_time_pattern}, \ref{table:twitter} and \ref{table:msmarco}.

Another important component of the proposed approach is the context aggregating multilayer perceptron (MLP) that receives the sequence representations of the two input passages and generates the scores, subsequently used to rank the passages. To evaluate its overall influence, the model is re-trained with a single linear layer replacing the entire multilayer perceptron that projects the sequence representation into a single scalar value that is passed into the loss function as the ranking score.

As seen in Table \ref{table:ablation}, while the approach still learns to rank the passages reasonably well without the context aggregating multilayer perceptron, the pair ranking accuracy significantly drops. This emphasises the importance of the MLP component to the overall predictive performance of our approach.

%............................................................................
\begin{table}[!t]
	\centering
	\resizebox{0.75\columnwidth}{!}{
		{\tabulinesep=0mm
			\begin{tabu}{@{\extracolsep{5pt}}l c @{}}
				\hline\hline
				\multicolumn{1}{l}{\multirow{2}{*}{Approach}} & 
				\multicolumn{1}{c}{Pair Labels}\T\B\\
				\cline{2-1}
						& Accuracy\T\B\\
				\hline\hline
Full Approach w/ ALBERT					& 0.861\T\\
Full Approach w/ RoBERTa				& 0.882\\
Full Approach w/ GPT2					& 0.899\B\\

\hline

Separate Transformers (BERT)			& 0.890\T\B\\

\hline

Approach w/o MLP (BERT)					& 0.815\T\B\\

\hline

Full Approach w/ BERT					& \textbf{0.902}\T\B\\

\hline
\hline
        \end{tabu}
    }
}
\captionsetup[table]{skip=7pt}
\captionof{table}{Pair-wise ranking accuracy results with different components of the proposed ranking approach removed or replaced, applied to 30\% of the Stack Exchange dataset \cite{StackExchange}.}
\label{table:ablation}
\end{table}
%............................................................................

Within our proposed pipeline, we opt for the use of a single Transformer model to produce the latent vector representing both passages in the input pair. However, one could envisage using two separate Transformer models, each independently learning the representation of one of the passages. While this will almost double the number of learnable parameters and can potentially introduce training instabilities, it is possible that the higher parameter count will enhance the learning process.

However, as seen in Table \ref{table:ablation}, the pair ranking accuracy is reduced when separate Transformer networks not sharing weights are used for the passages. This is primarily due to the reduction in the number of training samples for each Transformer and the possible overfitting of each network. Additionally, by using the same network for all passages, the model will get a better sense of the entire dataset and can produce more robust representations.
%--------------------------- END OF SUBSECTION ------------------------------
%-------------------------------------------------------------------------
%--------------------------- END OF SECTION ------------------------------
%-------------------------------------------------------------------------

%-------------------------------------------------------------------------
%----------------------------- NEW SECTION -------------------------------
%-------------------------------------------------------------------------
\section{Limitations and Future Work}
\label{sec:limits_future}

While ranking does not suffer from common issues that classification approaches suffer from, such as data imbalances, text subjectivity and lack of context, it does not suit every problem. Many text classification applications could be replaced by ranking with superior results, but there are certain situations where text classification remains the only option. For instance, a scenario wherein passages are meant to be categorised based on their topic cannot be addressed with ranking and requires a classification solution.

Moreover, text ranking cannot deal in absolutes and is only capable of relatively comparing items. For example, while ranking can provide an answer to whether a movie review is more or less positive than another, it cannot definitively say whether both the reviews are positive or negative. This issue could simply be addressed in real-world applications by identifying \enquote{sentinel} examples which are agreed, by consensus, to be positive / neutral / negative. Then using our ranking approach could indicate where an example is placed by comparing with the sentinels -- for example if it is ranked above a positive sentinel then it is a positive example. In this way, during inference, other passages can then be assigned an absolute labels/scores relative to their position in the sorted list produced by our ranking model based on the previously known sentinel passages.

It is important to note, however, that the inability to accurately provide absolute sentiments is seen in sentiment classification methods as well. Many classification solutions do not generalise to unseen data well and are incapable of absolute sentiment prediction if the vocabulary and linguistic structure of a new sentence has not been seen by the model within the training data.

Additionally, for our approach, we intend to address this in a future work with a double-headed model that simultaneously ranks and classifies the input passages. As the two heads can correct each other during training, better representation learning and thus more accurate results can be expected, as well as the ability to provide an absolute relevance and sentiment value for each passage.

Another avenue for future work would be to modify the loss function to cope with lists rather than pairs of passages. Whilst more difficult to optimise, list-wise ranking generally offers a better understanding of the context of the dataset, leading to more accurate results in addition to fewer post-processing requirements and thus better efficiency during inference.

%-------------------------------------------------------------------------
%--------------------------- END OF SECTION ------------------------------
%-------------------------------------------------------------------------

%-------------------------------------------------------------------------
%----------------------------- NEW SECTION -------------------------------
%-------------------------------------------------------------------------
\section{Conclusion}
\label{sec:conclusion}

We have investigated the applications of ranking in natural language processing using a novel pair-wise ranking approach and different publicly-available datasets. Our text ranking approach makes use of state-of-the-art Transformer networks to generate a learned representation of a pair of text sequences. These representation vectors are subsequently used as the input to a context-aggregating multilayer perceptron, which combines the features representing the context and the content of the passages, assesses the relationship between the two input passages and regresses to two values denoting the ranking scores subsequently used to rank the passages simply based on the ordering of the predicted scores.

The entirety of the model is trained end to end using a margin-based ranking loss function. Experiments are carried out on four publicly available datasets, with all the evaluation demonstrating the effectiveness of ranking for potential recommendation, forum moderation and security applications. We also compare the results of our approach directly with state-of-the-art classification methods. A comparison of our ranking results converted to class labels with state-of-the-art classification results indicates an approximately 22\% improvement, pointing to the efficacy of ranking over classification.
\\
\textit{We kindly invite the readers to refer to the supplemental \href{https://youtu.be/Njjg3mWVE2g}{\textbf{video}}: \url{https://youtu.be/Njjg3mWVE2g} for more information.}
%-------------------------------------------------------------------------
%--------------------------- END OF SECTION ------------------------------
%-------------------------------------------------------------------------

\bibliographystyle{IEEEtran}
\bibliography{custom}

\end{document}